\documentclass[runningheads]{llncs}
\usepackage[symbol]{footmisc}
\usepackage{graphicx}
\usepackage{amsmath,amssymb,amsfonts}
\usepackage{multirow}
\usepackage{algorithm}
\usepackage{algorithmicx}
\usepackage{algpseudocode}
\usepackage{hyperref}
\usepackage{multirow}

\newcommand{\set}[1]{\mathcal{#1}}

\DeclareMathOperator*{\loss}{{\ell}}

\DeclareMathOperator*{\argmin}{\arg\min}

\newcommand{\E}{\mathbb{E}}
\newcommand{\RN}{\mathbb{R}}

\DeclareMathOperator*{\regularization}{\ensuremath{{\theta}}}

\newcommand{\xorig}{\ensuremath{\vec{x}_{\text{orig}}}}
\newcommand{\yorig}{\ensuremath{y_{\text{orig}}}}

\newcommand{\setY}{\ensuremath{\set{Y}}}

\newcommand{\xcf}{\ensuremath{\vec{x}_{\text{cf}}}}

\newcommand{\ycf}{\ensuremath{y_{\text{cf}}}}
\newcommand{\CF}[2]{\ensuremath{\text{CF}(#1,#2)}}

\newcommand{\dimsym}{d}
\newcommand{\setD}{\set{D}}

\newcommand{\classifier}{\ensuremath{h}}

\newcommand{\myCF}[2]{\ensuremath{\text{CF}(#1,#2)}}

\newcommand{\refeq}[1]{Eq.~\eqref{#1}}
\newcommand{\refdef}[1]{Definition~\ref{#1}}

\begin{document}
\title{``Explain it in the Same Way!'' -- Model-Agnostic Group Fairness of Counterfactual Explanations\thanks{We gratefully acknowledge funding from the VW-Foundation for the project \textit{IMPACT} funded in the frame of the funding line \textit{AI and its Implications for Future Society}.}}
\titlerunning{Model-Agnostic Group Fairness of Counterfactuals}
%
\author{
Andr\'e Artelt \inst{1,2} \and
Barbara Hammer \inst{1}}

\authorrunning{A. Artelt and B. Hammer}
%
\institute{Bielefeld University, Bielefeld, Germany \and
University of Cyprus, Nicosia, Cyprus}
\maketitle              
\begin{abstract}
Counterfactual explanations are a popular type of explanation for making the outcomes of a decision making system transparent to the user. Counterfactual explanations tell the user what to do in order to change the outcome of the system in a desirable way. However, it was recently discovered that the recommendations of what to do can differ significantly in their complexity between protected groups of individuals. Providing more difficult recommendations of actions to one group leads to a disadvantage of this group compared to other groups.

In this work we propose a model-agnostic method for computing counterfactual explanations that do not differ significantly in their complexity between protected groups.

\end{abstract}
\section{Introduction}
Transparency and fairness are essential building blocks of modern trustworthy decision making systems deployed in the real world~\cite{toreini2020relationship,mehrabi2021survey}.
Transparency is often realized by means of explanations -- i.e. explanations of the system's behavior are provided to the user~\cite{molnar2019}. A lot of different explanation methodologies have been developed by researchers establishing the field of eXplainable Artificial Intelligence (XAI)~\cite{molnar2019,arrieta2020explainable}. Developed explanation methodologies include feature relevance/importance methods~\cite{FeatureImportance} and examples based methods~\cite{CaseBasedReasoning}. Instances of popular example based methods are contrasting explanations like counterfactual explanations~\cite{counterfactualwachter,CounterfactualReviewChallenges} and prototypes \& criticisms~\cite{PrototypesCriticism} -- these methods use a set or a single example for explaining the behavior of the system.
The aspect of fairness~\cite{mehrabi2021survey} usually refers to the requirement that the output (i.e. predictions) of the systems must not discriminate single individuals or groups of individuals.
Individual fairness states that similar individuals must be treated similarly~\cite{FairnessThroughAwareness} -- i.e. the behavior or output of a fair system should not differ significantly between similar individuals. The key problem here is to come up with a suitable definition of similarity.
On the other hand, group fairness states that the system must not behave significantly different between individuals from different protected groups, which are created based on protected attributes such as gender, race, etc., -- i.e. the behavior/output of the system must be independent of the protected attribute~\cite{10.1145/3494672,mehrabi2021survey}. Here the challenges is to model independence of the protected attribute -- different modelings give rise to different fairness criteria which contradict each other~\cite{10.1145/3494672}.

Both aspects, transparency/explanations and fairness, have been extensively studied in the past~\cite{adadi2018peeking,rawal2021recent,arrieta2020explainable,10.1145/3287560.3287589,10.1145/3494672}. Only recently, the research community started to investigated the fairness of different explanation methodologies and discovered that fairness issues might exist there as well~\cite{balagopalan2022road,von2022fairness}. In particular, it was observed that provided explanations might differ significantly between similar individuals or between protected groups of individuals~\cite{gupta2019equalizing,slack2021counterfactual,artelt2021evaluating}. This becomes problematic especially in case of actionable recourse -- i.e. explanations that tell the user what to do in order to observe a different outcome. A common way of achieving actionable recourse is to provide recommendations as explanations such as counterfactual explanations -- as some recommendations are easier to full fill than others, significant differences in these recommendations across groups or similar individuals can therefore disadvantage certain individuals or groups of individuals.

\paragraph*{Our contributions}
We propose a novel model-agnostic method for computing counterfactual explanations, a very popular instance of contrasting explanations, that satisfy a particular notion of group fairness -- i.e. the counterfactual explanations should not differ significantly between different protected groups. We also extensively empirically evaluate the performance of our proposed method in several experiments.

The remainder of this work is structured as follows: In Section~\ref{sec:counterfactuals} we review counterfactual explanations and (re-)introduce the group fairness problem of counterfactual explanations with which we are dealing in this work. Next, in Section~\ref{sec:contribution}, we propose our methodology for computing group fair counterfactual explanations, which we evaluate empirically in Section~\ref{sec:experiments}. Finally, after discussing important characteristics of our proposed method and its broader impact in Section~\ref{sec:discussion}, the work closes with a conclusion and summary in Section~\ref{sec:conclusion}.

\section{Counterfactual Explanations}\label{sec:counterfactuals}
Counterfactual explanations (often just called \textit{counterfactuals}) are an instance of contrasting explanations, stating changes to the input that lead to a different outcome (i.e. behavior of the system). Counterfactuals can be interpreted as a recommendation of actions that lead to a different specific behavior or output of the system. Since this mimics the way humans explain things~\cite{CounterfactualsHumanReasoning} -- i.e. humans often ask questions like \textit{``What needs to be changed in order to observe a different outcome?''} --, counterfactuals became quite popular in the XAI community~\cite{molnar2019,CounterfactualReviewChallenges}.
For illustrative purposes, consider the example of loan application: \textit{Imagine you applied for a credit at a bank. Unfortunately, the bank rejects your application. Now, you would like to know why. In particular, you would like to know what would have to be different so that your application would have been accepted. A possible explanation might be that you would have been accepted if you had earned $500$\$ more per month and if you had not had a second credit card.}
In order to keep the explanations ``simple'' -- i.e. easy to understand --, a common approach is to minimize the number of recommended actions by staying as close as possible to the original input. This is formalized in the following definition~\refdef{def:counterfactual}.
\begin{definition}[(Closest) Counterfactual Explanation~\cite{counterfactualwachter}]\label{def:counterfactual}
Assume a prediction function $\classifier:\RN^\dimsym \to \setY$ is given. Computing a counterfactual $\xcf \in \RN^\dimsym$ for a given input $\xorig \in \RN^\dimsym$ is phrased as the following optimization problem:
\begin{equation}\label{eq:counterfactualoptproblem}
\underset{\xcf \,\in\, \RN^\dimsym}{\arg\min}\; \loss\big(\classifier(\xcf), \ycf\big) + C \cdot \regularization(\xcf, \xorig)
\end{equation}
where $\loss(\cdot)$ denotes a loss function that penalizes deviation of the prediction $\classifier(\xcf)$ from the requested target prediction $\ycf$, $\regularization(\cdot)$ denotes a penalty for dissimilarity of $\xcf$ and $\xorig$\footnote{This acts as a regularization of the complexity of the final explanation -- i.e. prefer ``simple \& low complexity'' explanations.}, and $C>0$ denotes the regularization strength.
\end{definition}
Note that the actual explanation -- i.e. recommendation of actions to change some features in a particular way -- is given by the difference $\delta=\xcf-\xorig$. Furthermore, $\regularization(\cdot)$ can also be interpreted as the complexity of the counterfactual $\xcf$ -- i.e. measuring how difficult it is to execute the recommended changes $\delta$, since changing some features might be easier or more difficult than changing others, and also the amount of change might contribute to the complexity \& difficultiness of the explanation. For illustrative purposes, we again consider the previous example of a loan application: \textit{While a valid explanation might be to increase the monthly income by $500$\$, another valid explanation might be to increase monthly income by $400$\$, canceling a second credit card and having some property as a collateral security. Clearly, the latter one is much more difficult to full fill\footnote{Note that the notion of difficultiness and complexity of a given counterfactual highly depends on the domain and use-case -- in this work we use the 1-norm as a general proxy of the explanations complexity.} than the first one -- therefore, one would assign a higher score $\regularization(\cdot)$ to the latter.}

In the remainder of this work, we assume a binary classification problem: In this case, we denote a (closest) counterfactual $\xcf$ (according to~\refdef{def:counterfactual}) of a given sample $\xorig$ under a prediction function $\classifier(\cdot)$ simply as $\xcf=\CF{\xorig}{\classifier}$ and drop the target label $\ycf$ because it is uniquely determined.

The counterfactuals $\xcf$ from~\refdef{def:counterfactual} are also called \textit{closest counterfactuals}~\cite{karimi2021survey} because they try stay as close as possible to the original input $\xorig$. However, closest counterfactual are not necessarily plausible \& actionable which might hinder they use in practice~\cite{warren2022better} -- there exist numerous extensions~\cite{PlausibleCounterfactuals,CounterfactualGuidedByPrototypes,artelt2021convex,smyth2022few,poyiadzi2020face}, that try to improve plausibility \& actionability of counterfactuals from~\refdef{def:counterfactual}. The most simple approach for generating plausible, and therefore hopefully actionable, counterfactuals is to limit the feasible set of solutions to~\refeq{eq:counterfactualoptproblem} to a set of training samples~\cite{poyiadzi2020face} -- i.e. simply picking the most similar sample from the training set as a counterfactual explanation. While this option is easily implementable and guarantees plausible counterfactuals, it comes however at the price of potentially leaking sensitive information and might therefore pose a privacy issue~\cite{pawelczyk2022privacy}. Nevertheless, this is the method we use whenever we refer to plausible counterfactuals.

\subsection{Fairness of Counterfactual Explanations}
In this work, we assume that the individuals are grouped into two so called protected groups $\setD_1$ and $\setD_2$ based on some sensitive attribute like sex, race, etc.
While fairness of the predictions of AI systems is a well known challenge that has been studied for quite some time~\cite{mehrabi2021survey,pessach2022review,corbett2018measure}, potential fairness issues of explanations were only recently discovered by the community~\cite{von2022fairness,artelt2021evaluating,slack2021counterfactual,gupta2019equalizing}. In particular, it was observed that closest counterfactual explanations (\refdef{def:counterfactual}) might differ significantly between individuals from different groups~\cite{gupta2019equalizing} or also between similar individuals~\cite{artelt2021evaluating}.

In this work, we follow the common fairness notion of counterfactuals as introduced in~\cite{gupta2019equalizing}, that is based on the complexity \& difficultiness of the explanation, which is measured by the distance $\regularization(\cdot)$ between the original sample $\xorig$ and the corresponding counterfactual explanation $\xcf$ -- we refer to this quantity as the \textit{cost} of a counterfactual:
\begin{definition}[Cost of a Counterfactual Explanation]\label{def:cost_cf}
For a given sample $\xorig\in\RN^\dimsym$ and a classifier $\classifier(\cdot)$, we define the cost of the corresponding closest counterfactual~\refdef{def:counterfactual} $\xcf$ as the distance between the original sample and the counterfactual:
\begin{equation}
    \regularization(\xorig, \myCF{\xorig}{\classifier})
\end{equation}
where we substituted $\xcf$ with \myCF{\xorig}{\classifier} to stress the dependencies on $\xorig$, $\classifier(\cdot)$ and the explanation generation method $\myCF{\cdot}{\cdot}$ itself.
\end{definition}
Note that in~\refdef{def:cost_cf} we completely rely on $\regularization(\cdot)$ for measuring the complexity and costs of the recommended actions by the counterfactual explanation -- note that in practice, $\regularization(\cdot)$ must be chosen carefully and might require some domain knowledge.

It can be perceived as problematic if the cost of the counterfactuals (i.e. recommended actions) differs significantly between protected groups of individuals (e.g. males vs. females)~\cite{gupta2019equalizing,balagopalan2022road,slack2021counterfactual} or between similar individuals~\cite{artelt2021evaluating}, since achieving the algorithmic recourse would be much easier for the individuals getting low-cost explanations than those getting high-cost (i.e. more difficult to execute) explanations -- this would disadvantage certain groups of individuals which violates social norms of fairness~\cite{fehr1999theory,loewenstein1989social}.
For illustrative purposes, we again come back to our running example of loan applications: \textit{For instance, it can be considered as unfair if the counterfactual explanations presented to males on average ask only for a minor increase in their monthly income while those presented to females ask for a major increase in their monthly income.}
We formalize this intuition of group fairness of counterfactual explanations by using the definition of a cost~(\refdef{def:cost_cf}) of a counterfactual:
\begin{definition}[Group Fair Counterfactual Explanation]\label{def:groupfairness_cf}
We say that counterfactual explanations are fair iff their costs (\refdef{def:cost_cf}) does not differ significantly between the different groups~\cite{gupta2019equalizing,slack2021counterfactual} -- i.e. the distributions of $\regularization(\cdot)$ are ``sufficiently similar'':
\begin{equation}\label{eq:fairness:cf_dist}
    p_{\xorig\sim\setD_1}\Big(\regularization(\xorig, \myCF{\xorig}{\classifier})\Big) \approx p_{\xorig\sim\setD_2}\Big(\regularization(\xorig, \myCF{\xorig}{\classifier})\Big)
\end{equation}
\end{definition}
Since dealing with the entire distribution $p_{\xorig\sim\setD_?}$ might be too complicated, we relax and reduce~\refeq{eq:fairness:cf_dist} to the expected value~\cite{slack2021counterfactual}:
\begin{equation}\label{eq:fairness:cf_dist:relaxed}
    \underset{\xorig\sim\setD_1}{\E}\big[\regularization(\xorig, \myCF{\xorig}{\classifier})\big] \approx \underset{\xorig\sim\setD_2}{\E}\big[\regularization(\xorig, \myCF{\xorig}{\classifier})\big]
\end{equation}
which can be easily estimated using given sets of samples.

\paragraph{Existing work from literature}
In order to achieve group fairness of counterfactual explanations -- i.e. satisfying~\refeq{eq:fairness:cf_dist} -- in case of a support vector machine (SVM), the authors of~\cite{gupta2019equalizing} propose to change the training of the SVM such that the counterfactual explanations satisfy the group fairness criterion~\refeq{eq:fairness:cf_dist:relaxed}. More specifically, they propose an iterative method where they add recourse constraints to the original SVM optimization problem. In addition, they also propose a model-agnostic method where they iteratively generate samples using LIME~\cite{ribeiro2016model} that are used to retrain the original classifier $\classifier(\cdot)$ using these samples.

However, to the best of our knowledge, all existing methods either focus on specific models and/or do not consider changing the model-agnostic algorithm $\myCF{\cdot}{\cdot}$ for computing the final counterfactuals that satisfy~\refeq{eq:fairness:cf_dist}.
In this work, we aim to address this research gap by proposing a model-agnostic algorithm $\myCF{\cdot}{\cdot}$ for computing group fair counterfactual explanations (see Section~\ref{sec:contribution}) without retraining or changing the given model $\classifier(\cdot)$.

\section{Computing Group Fair Counterfactual Explanations}\label{sec:contribution}
In order to achieve the relaxed group fairness of counterfactual explanations (\refdef{def:groupfairness_cf}) as stated in~\refeq{eq:fairness:cf_dist:relaxed}, we propose to change the method $\myCF{\cdot}{\cdot}$ for computing counterfactuals. More specifically, we propose to extend~\refeq{eq:counterfactualoptproblem} as follows:
\begin{equation}\label{eq:counterfactualoptproblem:fair}
\begin{split}
    &\underset{\xcf\,\in\,\RN^\dimsym}{\min}\,\regularization(\xorig, \xcf) + C_0 \cdot \loss\left(\classifier(\xcf), \ycf\right) + C_1\cdot\max\left(z - \regularization(\xorig, \xcf)\right)\\
    &\text{where } z\sim p(\regularization(\vec{X}_\text{orig}, \vec{X}_\text{cf}))
\end{split}
\end{equation}
where the only difference to~\refeq{eq:counterfactualoptproblem} is the addition of the last regularization term $C_1\cdot\max\left(z - \regularization(\xorig, \xcf)\right)$ in~\refeq{eq:counterfactualoptproblem:fair} which adds a penalty if the distance (i.e. cost/complexity) of the counterfactual $\xcf$ is larger than some particular value $z$ sampled from the distribution of distances (i.e. costs) $\regularization(\cdot)$ of the \textit{disadvantaged group} -- by this we make sure that the difference between the advantaged and disadvantage group disappears on average.
Note that for this, we need access to a set of samples (including their counterfactual explanation) from the disadvantaged group, so that we can estimate $\E\big[\regularization(\vec{X}_\text{orig}, \vec{X}_\text{cf})\big]$ since we do not know the true distribution. For practical purposes, we suggest to simply select the distance (i.e. cost) of a random sample from the disadvantaged group as a value of $z$ -- by this we obtain a randomized algorithm/method for computing counterfactual explanations and the counterfactuals are no longer deterministic.
After we obtained a value of $z$, we can solve~\refeq{eq:counterfactualoptproblem:fair} either by using any black-box solver, like for instance Downhill-Simplex, resulting in a closest counterfactual or we could obtain a plausible counterfactual by simply select a sample from the training set $\setD_1 \cup \setD_2$ that minimizes~\refeq{eq:counterfactualoptproblem:fair}.
The entire proposed methodology for computing group fair counterfactual explanations is described in full detail in Algorithm~\ref{algo:fair_cf}.
\begin{algorithm}[t]
\caption{Computing Group Fair Counterfactual Explanations}\label{algo:fair_cf}
\textbf{Input:} A classifier $\classifier(\cdot)$; Samples grouped into two groups $\setD_1$ and $\setD_2$ based on their binary sensitive attribute; Regularization strengths $C_0, C_1>0$\\
\textbf{Output:} Group fair counterfactuals $\xcf$
\begin{algorithmic}[1]
 \State \Comment{\textbf{Initialization step}}
 \State \Comment{Compute the cost of counterfactual explanation for all samples in both sets}
 \State $\set{Z}_1 = \{\regularization(\xorig, \myCF{\xorig}{\classifier}) \quad \forall \xorig \in \setD_1\}$ 
 \State $\set{Z}_2 = \{\regularization(\xorig, \myCF{\xorig}{\classifier}) \quad \forall \xorig \in \setD_2\}$
 \State \Comment{Check which of the two groups is disadvantaged compared to the other}
 \State $\set{Z}=\emptyset$
 \If{$\frac{1}{|\set{Z}_1|}\sum_{z_i\in\set{Z}_1}z_i > \frac{1}{|\set{Z}_2|}\sum_{z_j\in\set{Z}_2}z_j$}
    \State $\set{Z}=\set{Z}_1$
 \Else
    \State $\set{Z}=\set{Z}_2$
 \EndIf
 \State \State \Comment{\textbf{Main step}}
 \State \Comment{For all new samples $(\xorig, \yorig)$ we compute a group fair counterfactual as follows:}
 \State $z=\textsc{random.choice}(\set{Z})$  \Comment{Select a random cost from the disadvantaged group}
 \State \Comment{Compute the group fair counterfactual by solving~\refeq{eq:counterfactualoptproblem:fair}}
 \State $\xcf = \underset{\xcf\,\in\,\RN^\dimsym}{\argmin}\,\regularization(\xorig, \xcf) + C_0 \cdot \loss\left(\classifier(\xcf), \ycf\right) + C_1\cdot\max\left(z - \regularization(\xorig, \xcf)\right)$
\end{algorithmic}
\end{algorithm}

Note that the major advantage of this approach, compared to existing methods~\cite{gupta2019equalizing}, is that the model $\classifier(\cdot)$ remains untouched -- i.e. we do not assume that we have access to model internals, or more crucially, that we are able to retrain the entire model which might be quite costly in practice -- which perfectly aligns with the model-agnostic paradigm of counterfactual explanations~\cite{counterfactualwachter}.

\section{Experiments}\label{sec:experiments}
We empirically evaluate our proposed method (Algorithm~\ref{algo:fair_cf}) for generating group fair counterfactual explanations on a diverse set of data sets and classifiers. We also consider and compare our proposed method (Algorithm~\ref{algo:fair_cf}) for two different types of counterfactuals: Closest counterfactuals (\refdef{def:counterfactual}) and plausible counterfactuals (i.e. samples from the training set). The Python implementation, including all necessary details for reproduction, of the experiments is publicly available on GitHub\footnote{\url{https://github.com/HammerLabML/ModelAgnosticGroupFairnessCounterfactuals}}.

\subsection{Data}
\paragraph{Credit card clients data set}
The Credit card clients data set~\cite{yeh2009comparisons} (\textit{Credit}) contains $30000$ data records of customers used for default payment prediction. Each record is described by $23$ attributes ($8$ categorical, $14$ numerical and $1$ binary). The sensitive attribute is ``sex'' but we also remove ``age'' and ``marriage''.

\paragraph{Communities \& Crime data set}
The Communities \& Crime data set~\cite{dheeru2017uci} (\textit{Crime}) contains $1994$ socio-economic data records from $46$ states in the USA. Following the preprocessing as suggested in~\cite{le2022survey}, we are left with $100$ encoded attributes and that are used to predict the crime rate (low crime rate vs. high crime rate). The sensitive attribute is ``race''.

\paragraph{Law school data set}
The Law school data set~\cite{wightman1998lsac} (\textit{Law}) contains $20798$ law school admission records. Each record (student) is described by $12$ attributes ($3$ categorical, $3$ binary and $6$ numerical). The binary target attribute describes whether the student as admitted or not. The sensitive attribute is ``sex'' but we also remove the attribute ``race''.

\subsection{Setup}
We run the following procedure, for every data set and classifier -- we consider logistic regression (\textit{Logreg}), decision trees (\textit{Dectree}) and Gaussian naive Bayes (\textit{GNB}) --, in a $3$-fold cross validation:
\begin{enumerate}
\item Build the classifier using all available training data -- i.e. ignore any group assignments based on sensitive attributes -- but sensitive attributes are NOT included as features.
\item Compute two counterfactual explanations for each sample of all test samples that are classified correctly -- the closest counterfactual (\refdef{def:counterfactual}) is computed by using~\cite{ceml} and for a plausible counterfactual we simply select the sample from the training set that minimizes~\refeq{eq:counterfactualoptproblem}. We then group these samples and their counterfactuals according to the sensitive attribute.
\item Use Algorithm~\ref{algo:fair_cf} to compute group fair counterfactuals of all correctly classified samples. Again, we solve~\refeq{eq:counterfactualoptproblem:fair} in two different ways: A closest counterfactual by solving~\refeq{eq:counterfactualoptproblem:fair} using the Downhill-Simplex method, and a plausible counterfactual by limiting the set of feasible solutions of~\refeq{eq:counterfactualoptproblem:fair} to the training set.
\end{enumerate}
For the purpose of evaluating our proposed method Algorithm~\ref{algo:fair_cf}, we compare the costs per group of the closest counterfactuals (\refdef{def:counterfactual}) with those of the group fair counterfactuals (\refdef{def:groupfairness_cf}).

\subsection{Results}
\begin{table}
\caption{Results for closest counterfactuals -- we report the median cost of counterfactuals (\refdef{def:cost_cf}), all number are rounded to two decimal points.}
\centering
\begin{tabular}{|c|c||c|c||c|c||c|c||}
 \hline
 Clf & \textit{Data set} & Group-0 & Group-1 & Group-0-Fair & Group-1-Fair & Diff & Diff-Fair  \\
 \hline\hline
  \multirow{3}{*}{\rotatebox[origin=c]{90}{Logreg}}
  & Credit & $0.96$ & $0.97$ & $1.62$ & $1.65$ & $0.01$ & $0.03$ \\
  & Crime & $1.75$ & $1.2$ & $2.06$ & $1.91$ & $0.55$ & $0.15$ \\
  & Law & $1.8$ & $1.81$ & $2.6$ & $2.63$ & $0.01$ & $0.02$ \\
 \hline
\multirow{3}{*}{\rotatebox[origin=c]{90}{Dectree}}
  & Credit & $0.23$ & $0.25$ & $0.43$ & $0.43$ & $0.02$ & $0.0$ \\
  & Crime & $1.52$ & $1.06$ & $1.61$ & $1.52$ & $0.46$ & $0.09$ \\
  & Law & $0.46$ & $0.45$ & $0.76$ & $0.75$ & $0.01$ & $0.01$ \\
 \hline
 \multirow{3}{*}{\rotatebox[origin=c]{90}{GNB}}
  & Credit & $6.71$ & $6.39$ & $10.26$ & $10.01$ & $0.32$ & $0.24$ \\
  & Crime & $7.06$ & $5.47$ & $7.21$ & $8.16$ & $1.59$ & $0.95$ \\
  & Law & $5.71$ & $5.73$ & $6.93$ & $6.95$ & $0.02$ & $0.02$ \\
 \hline
 \hline
\end{tabular}
\label{table:exp:results:closestCF}
\end{table}
\begin{table}
\caption{Results for plausible counterfactuals -- we report the median cost of counterfactuals (\refdef{def:cost_cf}), all number are rounded to two decimal points.}
\centering
\begin{tabular}{|c|c||c|c||c|c||c|c||}
 \hline
 Clf & \textit{Data set} & Group-0 & Group-1 & Group-0-Fair & Group-1-Fair & Diff & Diff-Fair  \\
 \hline\hline
  \multirow{3}{*}{\rotatebox[origin=c]{90}{Logreg}}
  & Credit & $3.81$ & $3.65$ & $10.33$ & $10.41$ & $0.17$ & $0.08$ \\
  & Crime & $62.1$ & $55.34$ & $61.86$ & $61.62$ & $6.76$ & $0.24$ \\
  & Law & $4.25$ & $4.31$ & $5.30$ & $5.33$ & $0.06$ & $0.03$ \\
 \hline
\multirow{3}{*}{\rotatebox[origin=c]{90}{Dectree}}
  & Credit & $3.03$ & $2.96$ & $10.13$ & $10.15$ & $0.07$ & $0.02$ \\
  & Crime & $63.04$ & $53.92$ & $61.73$ & $62.32$ & $9.12$ & $0.59$ \\
  & Law & $2.89$ & $2.97$ & $6.06$ & $6.41$ & $0.08$ & $0.35$ \\
 \hline
  \multirow{3}{*}{\rotatebox[origin=c]{90}{GNB}}
  & Credit & $4.47$ & $4.18$ & $9.87$ & $10.14$ & $0.28$ & $0.26$ \\
  & Crime & $74.42$ & $60.36$ & $58.88$ & $61.06$ & $14.05$ & $2.18$ \\
  & Law & $5.11$ & $5.16$ & $4.63$ & $4.63$ & $0.05$ & $0.01$ \\
 \hline
 \hline
\end{tabular}
\label{table:exp:results:plausibleCF}
\end{table}
A summary of results for the closest counterfactuals are shown in Table~\ref{table:exp:results:closestCF} and the results for plausible counterfactuals are shown in Table~\ref{table:exp:results:plausibleCF}.
\begin{figure}
    \centering
    \includegraphics[scale=.7]{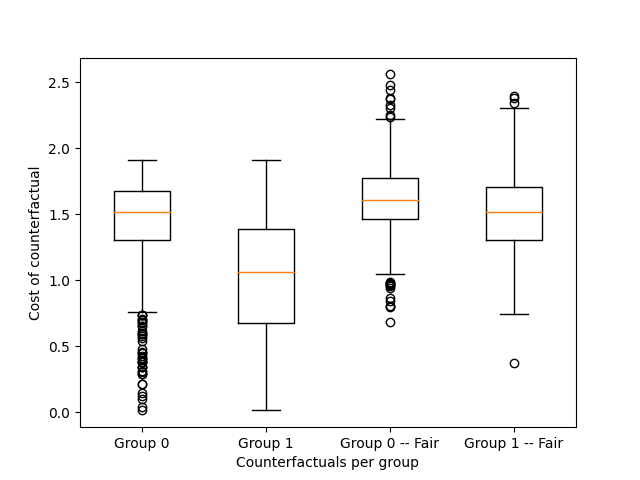}
    \caption{Normal vs. group fairness constrained counterfactuals: Group wise cost of closest counterfactuals for a decision tree classifier on the Communities \& Crime data set.}
    \label{fig:exp:resuls:cherrypick}
\end{figure}
We observe two things: First, not all scenarios show group fairness issues -- i.e. in some cases there is no significant difference in the costs across the different groups. Second, if there is a significant difference, our method almost always improves the group fairness significantly -- i.e. the difference in the costs across different groups is much smaller, which we additionally illustrate in Figure~\ref{fig:exp:resuls:cherrypick} for the particular scenario of closest counterfactuals of a decision tree classifier on the CommunitiesCrime data set. We do not observer any difference between closest and plausible counterfactuals.
These findings demonstrate the strong performance of our proposed model-agnostic method for computing group fair counterfactual explanations.

\section{Discussion \& Broader Impact}\label{sec:discussion}
\textbf{Advantages.} The key advantage of our proposed method over existing work is that it is completely model-agnostic -- i.e. we do not assume access to any model internals or that we are able to retrain the model.

\textbf{Computational complexity.} Before we can compute group fair counterfactuals, we first have to compute counterfactual explanations of all samples from the training set that is used to estimate the distribution of costs. The computational complexity of this step mainly depends on the number of samples and the classifier $\classifier(\cdot)$, since the computational complexity of computing a counterfactual varies significantly between different classifiers and how many assumptions we can make on the classifier~\cite{CounterfactualComputationSurvey}. However, this initial step has to be done only once and does not need to be repeated if the data distribution and model do not change.
The computation of the group fair counterfactual is then done by solving~\refeq{eq:counterfactualoptproblem:fair} using a black-box method like Downhill-Simplex method. 

\textbf{Limitations.} There exist two potential limitations of our proposed method:
First, the initialization step of Algorithm~\ref{algo:fair_cf} assumes access to a sufficiently large set of training samples that can be used to estimate the costs of counterfactuals across the two protected groups. In practice, it might be challenging to obtain enough samples from a minority group in order to reliable estimate the cost of the counterfactuals. Furthermore, the computation of these counterfactuals induces some computational costs as discussed above.
Second, the advantage of being model-agnostic, and not considering any special cases where we might have access to model internals, comes at the price that we can not exploit any structure of $\classifier(\cdot)$ for efficiently computing a group fair counterfactual by solving~\refeq{eq:counterfactualoptproblem:fair}. As a consequence, we have to rely on computationally expensive methods for solving~\refeq{eq:counterfactualoptproblem:fair}. Note that this is in contrast to the initialization step of Algorithm~\ref{algo:fair_cf}, where we have to compute normal counterfactuals and can exploit any structure or access to model internals if those are available.

\textbf{Broader impact.} Counterfactual explanations are widely used in practice to provide users with actionable feedback how to change the outcome of a decision making system. Potential fairness issues, as described in this work, are therefore harmful to individuals and must be eliminated as much as possible. We therefore expect that our proposed method have a high impact in practice as it can significantly improve the fairness of counterfactual explanations. Because it is completely model-agnostic, it can be applied to any models used by practitioners, and our publicly available Python implementation of Algorithm~\ref{algo:fair_cf} is easy to use and therefore allows a direct and seamless integration into existing XAI pipelines.

\section{Summary \& Conclusion}\label{sec:conclusion}
In this paper, we looked into the issue of group fairness of counterfactual explanations, and proposed a model-agnostic method for computing group fair counterfactuals. We empirically evaluated our proposed algorithm on a diverse set of experiments and observed a strong performance.
The key advantage over existing methods from literature is that our method does not need access to the model or have to retrain the model -- i.e. our method is completely model-agnostic and can be applied to any model that is going to be explained by counterfactuals.

Based on this work, there emerge a couple of potential directions for future research:
\begin{itemize}
    \item In this work we ignored any potential access to model internals or structure of $\classifier(\cdot)$ for computing group fair counterfactuals. However, as it was shown in~\cite{CounterfactualComputationSurvey}, exploiting model internals can lead to much more efficient algorithms for computing normal closest counterfactuals (\refdef{def:counterfactual}). In this context it would interesting and relevant to study how such available knowledge could be exploited for computing group fair counterfactuals by solving~\refeq{eq:counterfactualoptproblem:fair}.
    \item As discussed, the initialization step of Algorithm~\ref{algo:fair_cf} has to be recomputed if the data distribution or $\classifier(\cdot)$ changes. Since the initialization step can be computationally expensive, it would be relevant to study if and how an adaptation of this initialization step is possible in case of any changes in the data distribution or the model $\classifier(\cdot)$.
\end{itemize}

%
%
%
\bibliographystyle{splncs04}
\bibliography{mybibliography}

\begin{thebibliography}{10}
\providecommand{\url}[1]{\texttt{#1}}
\providecommand{\urlprefix}{URL }
\providecommand{\doi}[1]{https://doi.org/#1}

\bibitem{CaseBasedReasoning}
Aamodt, A., Plaza., E.: Case-based reasoning: Foundational issues,
  methodological variations, and systemapproaches. AI communications  (1994)

\bibitem{adadi2018peeking}
Adadi, A., Berrada, M.: Peeking inside the black-box: a survey on explainable
  artificial intelligence (xai). IEEE access  \textbf{6},  52138--52160 (2018)

\bibitem{arrieta2020explainable}
Arrieta, A.B., D{\'\i}az-Rodr{\'\i}guez, N., Del~Ser, J., Bennetot, A., Tabik,
  S., Barbado, A., Garc{\'\i}a, S., Gil-L{\'o}pez, S., Molina, D., Benjamins,
  R., et~al.: Explainable artificial intelligence (xai): Concepts, taxonomies,
  opportunities and challenges toward responsible ai. Information fusion
  \textbf{58},  82--115 (2020)

\bibitem{CounterfactualComputationSurvey}
Artelt, A., Hammer, B.: On the computation of counterfactual explanations - {A}
  survey. CoRR  \textbf{abs/1911.07749} (2019)

\bibitem{PlausibleCounterfactuals}
Artelt, A., Hammer, B.: Convex density constraints for computing plausible
  counterfactual explanations. 29th International Conference on Artificial
  Neural Networks (ICANN) (2020)

\bibitem{artelt2021convex}
Artelt, A., Hammer, B.: Convex optimization for actionable$\backslash$\&
  plausible counterfactual explanations. arXiv preprint arXiv:2105.07630
  (2021)

\bibitem{artelt2021evaluating}
Artelt, A., Vaquet, V., Velioglu, R., Hinder, F., Brinkrolf, J., Schilling, M.,
  Hammer, B.: Evaluating robustness of counterfactual explanations. In: 2021
  IEEE Symposium Series on Computational Intelligence (SSCI). pp. 01--09. IEEE
  (2021)

\bibitem{ceml}
Artelt, A.: Ceml: Counterfactuals for explaining machine learning models - a
  python toolbox. \url{https://www.github.com/andreArtelt/ceml} (2019 - 2022)

\bibitem{balagopalan2022road}
Balagopalan, A., Zhang, H., Hamidieh, K., Hartvigsen, T., Rudzicz, F.,
  Ghassemi, M.: The road to explainability is paved with bias: Measuring the
  fairness of explanations. arXiv preprint arXiv:2205.03295  (2022)

\bibitem{CounterfactualsHumanReasoning}
Byrne, R.M.J.: Counterfactuals in explainable artificial intelligence (xai):
  Evidence from human reasoning. In: {IJCAI-19} (2019)

\bibitem{corbett2018measure}
Corbett-Davies, S., Goel, S.: The measure and mismeasure of fairness: A
  critical review of fair machine learning. arXiv preprint arXiv:1808.00023
  (2018)

\bibitem{dheeru2017uci}
Dheeru, D., Taniskidou, E.K.: Uci machine learning repository  (2017)

\bibitem{FairnessThroughAwareness}
Dwork, C., Hardt, M., Pitassi, T., Reingold, O., Zemel, R.S.: Fairness through
  awareness. In: Goldwasser, S. (ed.) Innovations in Theoretical Computer
  Science 2012, Cambridge, MA, USA, January 8-10, 2012. pp. 214--226. {ACM}
  (2012). \doi{10.1145/2090236.2090255},
  \url{https://doi.org/10.1145/2090236.2090255}

\bibitem{fehr1999theory}
Fehr, E., Schmidt, K.M.: A theory of fairness, competition, and cooperation.
  The quarterly journal of economics  \textbf{114}(3),  817--868 (1999)

\bibitem{FeatureImportance}
{Fisher}, A., {Rudin}, C., {Dominici}, F.: {All Models are Wrong but many are
  Useful: Variable Importance for Black-Box, Proprietary, or Misspecified
  Prediction Models, using Model Class Reliance}. arXiv e-prints
  arXiv:1801.01489 (Jan 2018)

\bibitem{10.1145/3287560.3287589}
Friedler, S.A., Scheidegger, C., Venkatasubramanian, S., Choudhary, S.,
  Hamilton, E.P., Roth, D.: A comparative study of fairness-enhancing
  interventions in machine learning. p. 329–338. FAT* '19, Association for
  Computing Machinery, New York, NY, USA (2019). \doi{10.1145/3287560.3287589},
  \url{https://doi.org/10.1145/3287560.3287589}

\bibitem{gupta2019equalizing}
Gupta, V., Nokhiz, P., Roy, C.D., Venkatasubramanian, S.: Equalizing recourse
  across groups. arXiv preprint arXiv:1909.03166  (2019)

\bibitem{karimi2021survey}
Karimi, A.H., Barthe, G., Sch{\"o}lkopf, B., Valera, I.: A survey of
  algorithmic recourse: contrastive explanations and consequential
  recommendations. ACM Computing Surveys (CSUR)  (2021)

\bibitem{PrototypesCriticism}
Kim, B., Koyejo, O., Khanna, R.: Examples are not enough, learn to criticize!
  criticism for interpretability. In: Advances in Neural Information Processing
  Systems 29 (2016)

\bibitem{le2022survey}
Le~Quy, T., Roy, A., Iosifidis, V., Zhang, W., Ntoutsi, E.: A survey on
  datasets for fairness-aware machine learning. Wiley Interdisciplinary
  Reviews: Data Mining and Knowledge Discovery p. e1452 (2022)

\bibitem{loewenstein1989social}
Loewenstein, G.F., Thompson, L., Bazerman, M.H.: Social utility and decision
  making in interpersonal contexts. Journal of Personality and Social
  psychology  \textbf{57}(3), ~426 (1989)

\bibitem{CounterfactualGuidedByPrototypes}
Looveren, A.V., Klaise, J.: Interpretable counterfactual explanations guided by
  prototypes pp. 650--665 (2021)

\bibitem{mehrabi2021survey}
Mehrabi, N., Morstatter, F., Saxena, N., Lerman, K., Galstyan, A.: A survey on
  bias and fairness in machine learning. ACM Computing Surveys (CSUR)
  \textbf{54}(6),  1--35 (2021)

\bibitem{molnar2019}
Molnar, C.: Interpretable Machine Learning (2019)

\bibitem{pawelczyk2022privacy}
Pawelczyk, M., Lakkaraju, H., Neel, S.: On the privacy risks of algorithmic
  recourse. arXiv preprint arXiv:2211.05427  (2022)

\bibitem{10.1145/3494672}
Pessach, D., Shmueli, E.: A review on fairness in machine learning. ACM Comput.
  Surv.  \textbf{55}(3) (feb 2022). \doi{10.1145/3494672},
  \url{https://doi.org/10.1145/3494672}

\bibitem{pessach2022review}
Pessach, D., Shmueli, E.: A review on fairness in machine learning. ACM
  Computing Surveys (CSUR)  \textbf{55}(3),  1--44 (2022)

\bibitem{poyiadzi2020face}
Poyiadzi, R., Sokol, K., Santos-Rodriguez, R., De~Bie, T., Flach, P.: Face:
  feasible and actionable counterfactual explanations. In: Proceedings of the
  AAAI/ACM Conference on AI, Ethics, and Society. pp. 344--350 (2020)

\bibitem{rawal2021recent}
Rawal, A., Mccoy, J., Rawat, D.B., Sadler, B., Amant, R.: Recent advances in
  trustworthy explainable artificial intelligence: Status, challenges and
  perspectives. IEEE Transactions on Artificial Intelligence  \textbf{1}(01),
  ~1--1 (2021)

\bibitem{ribeiro2016model}
Ribeiro, M.T., Singh, S., Guestrin, C.: Model-agnostic interpretability of
  machine learning. arXiv preprint arXiv:1606.05386  (2016)

\bibitem{slack2021counterfactual}
Slack, D., Hilgard, A., Lakkaraju, H., Singh, S.: Counterfactual explanations
  can be manipulated. Advances in Neural Information Processing Systems
  \textbf{34},  62--75 (2021)

\bibitem{smyth2022few}
Smyth, B., Keane, M.T.: A few good counterfactuals: generating interpretable,
  plausible and diverse counterfactual explanations. In: International
  Conference on Case-Based Reasoning. pp. 18--32. Springer (2022)

\bibitem{toreini2020relationship}
Toreini, E., Aitken, M., Coopamootoo, K., Elliott, K., Zelaya, C.G.,
  Van~Moorsel, A.: The relationship between trust in ai and trustworthy machine
  learning technologies. In: Proceedings of the 2020 conference on fairness,
  accountability, and transparency. pp. 272--283 (2020)

\bibitem{CounterfactualReviewChallenges}
Verma, S., Dickerson, J., Hines, K.: Counterfactual explanations for machine
  learning: A review (2020)

\bibitem{von2022fairness}
Von~K{\"u}gelgen, J., Karimi, A.H., Bhatt, U., Valera, I., Weller, A.,
  Sch{\"o}lkopf, B.: On the fairness of causal algorithmic recourse. In:
  Proceedings of the AAAI Conference on Artificial Intelligence. vol.~36, pp.
  9584--9594 (2022)

\bibitem{counterfactualwachter}
Wachter, S., Mittelstadt, B., Russell, C.: Counterfactual explanations without
  opening the black box: Automated decisions and the gdpr. Harv. JL \& Tech.
  \textbf{31}, ~841 (2017)

\bibitem{warren2022better}
Warren, G., Smyth, B., Keane, M.T.: “better” counterfactuals, ones people
  can understand: Psychologically-plausible case-based counterfactuals using
  categorical features for explainable ai (xai). In: International Conference
  on Case-Based Reasoning. pp. 63--78. Springer (2022)

\bibitem{wightman1998lsac}
Wightman, L.F.: Lsac national longitudinal bar passage study. lsac research
  report series.  (1998)

\bibitem{yeh2009comparisons}
Yeh, I.C., Lien, C.h.: The comparisons of data mining techniques for the
  predictive accuracy of probability of default of credit card clients. Expert
  systems with applications  \textbf{36}(2),  2473--2480 (2009)

\end{thebibliography}

\end{document}